%% file: root.tex
\documentclass[letterpaper, 10 pt, conference]{ieeeconf}  

\IEEEoverridecommandlockouts                              

\usepackage{color}
\usepackage{multirow}
\usepackage{rotating}
\usepackage{gensymb}
\usepackage{wrapfig}
\usepackage{soul}
\usepackage{siunitx}
\usepackage{caption}
\usepackage{mathtools}
\usepackage{subcaption}
\usepackage[ruled]{algorithm2e} 

\usepackage[x11names]{xcolor} 
\usepackage{booktabs}

\definecolor{turquoise}{cmyk}{0.65,0,0.1,0.1}
\definecolor{purple}{rgb}{0.65,0,0.65}
\definecolor{darkgreen}{rgb}{0.0, 0.5, 0.0}
\definecolor{darkred}{rgb}{0.5, 0.0, 0.0}
\definecolor{darkblue}{rgb}{0.0, 0.0, 0.5}
\definecolor{blue}{rgb}{0.0, 0.0, 1.0}
\definecolor{orange}{rgb}{1.0, 0.5, 0.0}
\definecolor{red}{rgb}{1.0, 0.0, 0.0}
\definecolor{cherry}{RGB}{186,12,47}
\definecolor{lightblue_google}{RGB}{215,225,248}
\definecolor{blue_google}{RGB}{122,157,232}
\definecolor{red_google}{RGB}{238,103,92}
\definecolor{yellow_google}{RGB}{252,201,52}
\definecolor{green_google}{RGB}{91,185,116}
\definecolor{pink_google}{RGB}{161,66,244}
\definecolor{teal_google}{RGB}{0,172,193}

\usepackage{graphics}
\usepackage{pgf-pie}
\usepackage{pgfplots}
\pgfplotsset{compat=1.18}
\usepackage{tabularx}
\usepackage{cite}
\usepackage{placeins}
\usepackage{xspace}
\usepackage{svg}
\usepackage{pgfplots}
\pgfplotsset{compat=1.18}
\usepackage{multirow}
\usepackage{wrapfig}
\usepackage{makecell}

\def\papername{HUMEMBR\xspace}

\title{\LARGE \bf
\papername: Learning Human Routines for \\ Predictive Embodied Navigation
}
\author{Samira Huber$^{1}$, Klaas Pelzer$^{1}$, Duc M. Nguyen$^{2}$, Xuesu Xiao$^{2}$, S\"oren Pirk$^{1}$%
\thanks{$^{1}$Kiel University, Germany}%
\thanks{$^{2}$George Mason University, USA}%
}

\begin{document}
\bstctlcite{BSTcontrol}

\maketitle
\thispagestyle{empty}
\pagestyle{empty}

\input{content/00-abstract}
\input{content/01-introduction}
\input{content/02-related-work}
\input{content/03-method}
\input{content/04-experiments}

\input{content/05-results}

\input{content/privacy}
\input{content/06-conclusions}
\input{content/acknowledgement}



\end{document}

%% file: content/00-abstract.tex
\begin{abstract}
Understanding and navigating human-centered environments over extended periods of time while considering human behavior and routines remains a fundamental challenge in robotics. In real-world settings, robots may be asked to locate a specific individual, predict where that person is likely to be, or estimate when they typically leave a building. Addressing such queries requires reasoning over extensive histories of observations and capturing long-term behavioral patterns. To this end, we introduce Human-Centered Memory for Embodied Robots (\papername), a system designed for embodied question answering and routine-conditioned navigation. \papername integrates a continuous memory construction process with a parallel retrieval and querying mechanism, enabling the system to accumulate structured representations of human routines while supporting interactive, user-driven queries. Our experimental results indicate that \papername improves long-horizon reasoning about human behavior relative to full-context LLM baselines, while using substantially fewer tokens. Furthermore, we deploy \papername on a physical robot in two distinct environments, showing its ability to handle diverse queries and navigation tasks under real-world conditions. The code, videos, benchmark questions, and execution logs can be found at: https://samirahuber.github.io/humembr/
\end{abstract}

%% file: content/01-introduction.tex
\section{INTRODUCTION}
Robots are increasingly deployed in real-world environments such as offices and households, where they operate in dynamic settings shaped by human daily routines.
In such environments, the world is not purely spatial but structured by recurring behavioral patterns and temporal regularities: people arrive and leave at predictable times, occupy preferred locations, and interact with specific objects in habitual ways. For a robot to function effectively in these environments, it must reason not only about geometry and semantics, but also about human-centered temporal structure.

Conventional robotic representations, such as metric maps~\cite{Cadena_2016} or semantic scene graphs \cite{Armeni20193DSG,Werby-RSS-24, doi:10.1177/02783649211056674}, primarily encode static spatial information. While these representations are sufficient for obstacle avoidance and short-term navigation, they do not capture long-term patterns of human activity. Thus, current systems lack the ability to accumulate structured knowledge about recurring behaviors and leverage it for reasoning, question answering, or routine-aware navigation.

At the same time, recent advances in large~language~models~(LLMs) have enabled robots to interpret natural-language queries and decompose high-level instructions into actionable plans \cite{chiang_mobility_2024, ahn_as_2022, shah_lm-nav_2022, espada_leveraging_2025, bian_large_2025}. Embodied question answering and retrieval-augmented reasoning frameworks demonstrate that robots can access stored observations to answer queries about past events \cite{park2023generativeagentsinteractivesimulacra,ginting2025entermindpalacereasoning, anwar_remembr_2024}. However, existing approaches typically focus on object-centric reasoning or operate over limited temporal horizons. They do not explicitly model multi-day human routines, nor do they maintain persistent, identity-aware memory of dynamic human behavior.
This gap is particularly critical for long-term deployments. A robot operating daily in an office should be able to answer questions such as when a specific person typically arrives, where they are usually found at certain times, or how their routine changes over time. Enabling such capabilities requires persistent memory construction, identity tracking across days, and mechanisms for structured temporal retrieval - capabilities that are largely absent from current embodied systems.

\begin{figure}[t]
\centering
\includegraphics[width=\linewidth]{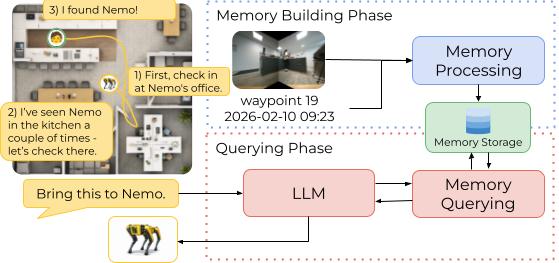}
\caption{\papername operates two parallel processes: Memory Building (blue) and Querying (red). During the Memory Building phase, the robot’s sensor data are recorded, images are captioned, and detected individuals are clustered. In the Querying phase, functions provide access to the stored information for robot navigation.}
\vspace{-4mm}
\label{fig:summary}
\end{figure}

We address these limitations by developing a human-centered, long-horizon memory framework for robot navigation and embodied question answering.
Our system enables robots to answer natural language queries and execute navigation tasks grounded in models of human daily routines.

Our approach maintains a continuous memory construction process alongside a retrieval and query module (see Fig.~\ref{fig:summary}).
This design allows the robot to incrementally gather information about human activities (``Memory Building Phase'') and to reason over long-term behavioral patterns in response to user queries (``Querying Phase'').
For example, the system can answer questions such as “When does Nemo usually arrive at work?” or execute commands such as “Search Nemo” by leveraging accumulated routine knowledge.
To evaluate long-horizon, human-centered reasoning, we introduce PersonEQA, the first benchmark for Person Embodied Question Answering with a focus on modeling human behavior and recurring routines. The benchmark comprises diverse human-centered queries spanning six answer categories and varying levels of complexity.

Through our experiments, we demonstrate that \papername effectively leverages long-horizon memory to perform human-centered reasoning over multi-day observations. Structured memory construction and targeted retrieval achieve higher accuracy on spatial, temporal, and person identification questions than a full-context baseline, while using substantially fewer tokens. We validate our approach across two settings via real-world deployment on a Boston Dynamics Spot robot \cite{bostondynamics_spot}, showing that accumulated memory supports navigation and person-centered tasks in a live office environment.

In summary, our contributions are that we: (1) design the PersonEQA benchmark to evaluate whether (a) clustering mechanisms correctly identify individuals over time and (b) an LLM can reason about human routines and interactions from accumulated memory; (2) introduce \papername, a retrieval-augmented LLM agent that performs function calls to access relevant memories and reason over them to answer human-centered questions; and (3) provide qualitative results from real-world deployments in two environments differing in scale, layout, and crowd density, demonstrating robust long-horizon reasoning and navigation.

Unlike prior retrieval-based embodied systems, our approach introduces an identity-aware, long-horizon memory (multiple days) that models recurring human routines rather than isolated observations. This enables predictive reasoning about human behavior (e.g., estimating likely locations or arrival times), which is not supported by existing object-centric or episodic memory frameworks.

%% file: content/02-related-work.tex
\section{Related Work}

\textbf{Embodied question answering (EQA).} EQA studies agents that acquire visual information through navigation in order to answer natural-language queries. Existing benchmarks primarily focus on object-centric reasoning in static environments, where agents infer object attributes, spatial relations, or scene states via exploration or episodic memory retrieval \cite{majumdar_openeqa_nodate, ren_explore_2024, zhao_cityeqa_2025}.

Human-centered understanding and long-horizon temporal reasoning remain largely underexplored. Current datasets either assume static scenes without persistent dynamic entities \cite{majumdar_openeqa_nodate, ren_explore_2024, zhao_cityeqa_2025}, or analyze human activities without embodied interaction and navigation capabilities \cite{grauman_ego4d_2022}. While Ginting et al. \cite{ginting2025entermindpalacereasoning} incorporate dynamic objects and long-horizon memory into embodied reasoning, their framework does not explicitly model persistent human identities, recurring routines, or daily behavioral patterns.

In contrast, we study embodied question answering grounded in dynamic human behavior, where agents must reason over multi-day observations and generate navigation goals based on accumulated spatial and temporal memory.

\textbf{Semantic Navigation.}
Traditional robotic navigation optimizes for metric goals, whereas human-centered environments require grounding semantic instructions in perceptual representations. Recent language-guided navigation approaches leverage LLMs and VLMs to interpret natural-language commands and retrieve relevant observations or prior experience \cite{zhou_navgpt_2023, chiang_mobility_2024, anwar_remembr_2024, yadav_findingdory_2025}. Although methods such as ReMEmbR \cite{anwar_remembr_2024} and Enter the Mind Palace \cite{ginting2025entermindpalacereasoning} incorporate long-term memory, they focus on retrospective state retrieval and do not model persistent human identities or recurring routines. In contrast, we formulate semantic navigation as routine-aware prediction, estimating likely locations from multi-day observations of human behavior.

\textbf{Vision-language models and robotics.} 
LLMs and VLMs are increasingly integrated into robotic systems for high-level reasoning, task planning, and instruction grounding. Prior work demonstrates that such models can translate natural-language commands into executable navigation policies \cite{chiang_mobility_2024, ahn_as_2022, shah_lm-nav_2022, espada_leveraging_2025, bian_large_2025}, and have also been applied to motion planning and manipulation by generating structured action sequences or parameterized control commands \cite{singh_progprompt_2022, sun_interactive_2023}. Despite these advances, most approaches address single-task execution or short-horizon planning and assume that relevant contextual information is available at query time. They do not explicitly construct persistent, structured memory representations of long-term environmental dynamics. In contrast, our work emphasizes long-horizon perception and memory construction to enable reasoning about dynamic human behavior across multiple days.

\textbf{Person Clustering and Re-Identification (ReID).} Long-term person perception in dynamic environments has been extensively studied, particularly in the context of person ReID and unsupervised clustering. Recent approaches integrate fine-grained feature disentanglement, cross-domain adaptation, and synthetic data augmentation to mitigate domain shift and long-term appearance variation \cite{Somers_2024, he2025instructreidmultipurposepersonreidentification, liang2025differdisentanglingidentityfeatures, pathak2025colorscolorsignoreclothes}.
Our work focuses on deployable long-horizon person association in a mobile robotic setting. We adopt a combination of clustering and ReID to increase robustness to appearance changes such as clothing and hairstyles, using face cues as a anchor when available, and body-based ReID features to recover associations when faces are missing or occluded.

Unlike existing systems, we address long-horizon, human-centered question answering. Our robot maintains persistent, identity-aware memory of human routines accumulated over extended periods, enabling queries that range from direct retrieval to multi-step reasoning over temporally distributed human behavior.

%% file: content/03-method.tex
\section{Methodology}

Robots deployed in real-world environments must not only operate under dynamically changing conditions (objects may move around) but they must also consider complex human behavior and recurring human routines. When performing daily tasks,  (e.g., cleaning a room, emptying a dishwasher, etc.) humans constantly change and manipulate their environment. 
Consequently, relevant information of humans, objects, and the environment itself may rapidly become outdated. In contrast to prior frameworks that separate memory construction from downstream reasoning \cite{anwar_remembr_2024, chiang_mobility_2024}, our method therefore maintains an up-to-date environmental representation by performing memory construction and memory querying concurrently in real time. This enables the system to ground decisions in recent observations while preserving long-term behavioral patterns accumulated over extended periods.

Specifically, our approach addresses two central challenges: (1) we aim to continuously construct a compact and structured memory that captures relevant information about individuals and their activities, enabling persistent tracking across multiple days despite variations in clothing, hairstyle, illumination, and viewpoint (see \ref{subsec:MemoryBuilding}); and (2) we query this memory to answer natural-language questions and solve person-centered tasks with a closed-loop framework that integrates reasoning, retrieval, and action (see~\ref{subsec:QuestionAnsweing}).

\begin{figure}[t]
\centering
\includegraphics[width=\linewidth]{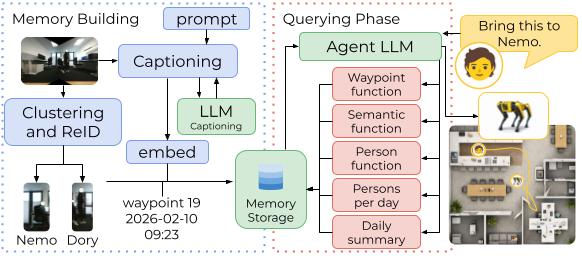}
\caption{\papername operates two parallel processes: Memory Building (left) and Querying (right). During the Memory Building phase, the robot’s sensor data are recorded, images are captioned, and detected individuals are clustered. In the Querying phase, up to five functions (red) are available to access the stored information (Memory Storage) and to control robot navigation.}
\vspace{-4mm}
\label{fig:methodolegy}
\end{figure}

\subsection{Memory Building}
\label{subsec:MemoryBuilding}
The memory construction module builds a structured long-term representation that goes beyond conventional scene mapping. In addition to storing scene descriptions, the system explicitly maintains a persistent, identity-aware memory of humans, enabling navigation conditioned on dynamic human behavior rather than solely on static spatial structure.

\begin{figure}[t]
\centering
\includegraphics[width=\linewidth]{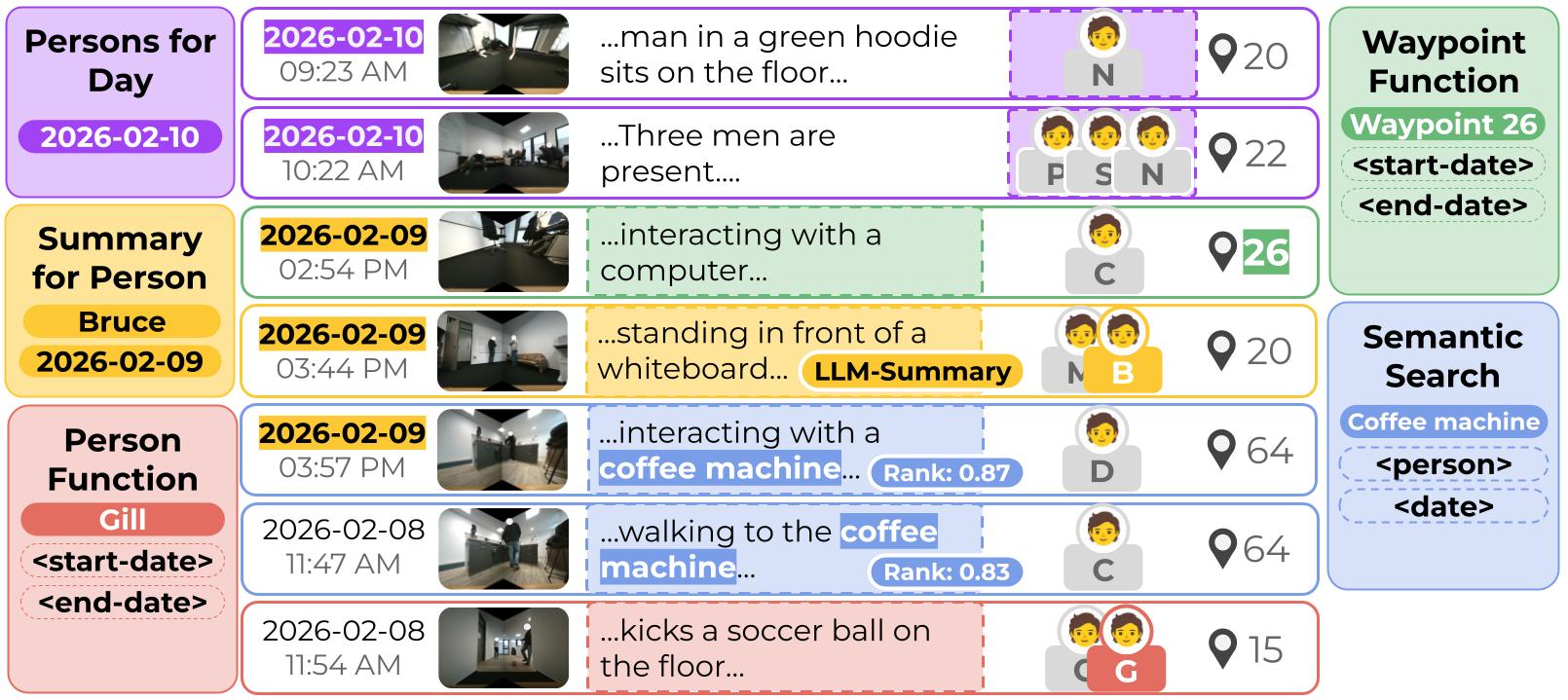}
\caption{The system exposes five retrieval functions (violet, yellow, red, green and blue) to access relevant observations. The dotted boxes denote the outputs returned to the LLM. For the Summary-for-Person function (yellow), all retrieved captions are forwarded to an LLM for aggregation prior to being provided to the agent. For Persons-for-Day (violet), only the extracted person identities are returned, rather than the full caption content. The dotted parameters are optional.}
\vspace{-4mm}
\label{fig:data_structure}
\end{figure}

While the robot captures data at 2~Hz, the incoming RGB observations are filtered using a ResNet50-based feature encoder \cite{he2016deep} to reduce redundant storage.  The visual embedding of each new frame is compared against that of the most recently retained frame via cosine similarity, and observations exceeding a predefined similarity threshold are discarded. This mechanism ensures retention of only visually distinctive frames, thereby enabling continuous acquisition while maintaining manageable memory growth.
Each retained memory entry consists of the RGB image, the corresponding spatial reference provided by the GraphNav system \cite{BD_GraphNavPythonClient}, and an associated timestamp. To incorporate semantic context, each image is annotated with a descriptive caption generated by Qwen-3-VL 235 B \cite{Qwen-VL}. The resulting caption is embedded into a dense vector representation using \textit{mxbai-embed-large-v1} \cite{emb2024mxbai}, enabling efficient similarity-based retrieval during the querying phase (see Fig.~\ref{fig:data_structure}).

Human instances are detected using a YOLO-based detector \cite{ultralytics_2024}, conditioned on the visibility of a sufficient number of body keypoints. Those detections extend the captions, since they do not provide persistent identity assignment for the observed individuals. For each detected individual, two complementary embedding models are applied to obtain discriminative identity representations.
Facial recognition serves as the primary identity anchor, as facial features remain invariant to clothing variations. We employ InsightFace \cite{insightface2023github} to extract facial embeddings whenever a face is visible. However, face observations are not consistently available in real-world settings, as individuals may be observed from the side or back during navigation, or their faces may be occluded due to viewpoint. To increase coverage and robustness, we integrate a second embedding model based on Keypoint Promptable Re-Identification (KPR) \cite{Somers_2024}. This model extracts full-body appearance features and remains effective under pose variation and partial occlusion. The combination of facial and full-body embeddings enables complementary identity cues across varying observation conditions.

Identity association is performed using an unsupervised two-stage clustering framework. In dynamic environments, individuals continuously enter and leave the scene, requiring incremental and non-parametric grouping. In the first stage, detections with available facial embeddings are clustered using an online DBSCAN algorithm \cite{ester1996dbscan}. Clustering is restricted to facial embeddings to establish high-confidence identity anchors while minimizing false merges.
In the second stage, detections without visible faces are associated with existing identity clusters using their ReID embeddings through nearest-neighbor matching under a predefined similarity threshold. The threshold balances two competing objectives: minimizing incorrect identity associations while maximizing assignment coverage.

\subsection{Question Answering and Semantic Navigation}
\label{subsec:QuestionAnsweing}
Given a natural-language query, the system must retrieve relevant information from long-term memory and translate the resulting reasoning into executable actions. For instance, in response to a query such as “Search Nemo,” the agent must infer previously observed locations associated with the specified individual, estimate the most probable current location based on historical patterns, and generate a sequence of candidate waypoints. 

Processing the complete memory with a transformer-based model is computationally inefficient and risks diluting relevant context. We therefore adopt a retrieval-augmented strategy that selects a query-conditioned subset of memory entries. Retrieval operates at multiple levels of abstraction: some functions access high-level summaries (e.g., day-level activity aggregation), while others retrieve fine-grained information such as individual observations, waypoint-specific logs, or temporally constrained person histories.

The language model interacts with memory through structured retrieval functions:

\begin{itemize}
\item \textbf{Semantic observations} $f(Q, P, t_e)$:  
Retrieves the top-n observations (with $n=30$) semantically similar to query $Q$ using embedding-based similarity search. 
We compute a relevance score following Park et al. \cite{park2023generativeagentsinteractivesimulacra} that combines semantic relevance with temporal recency
\begin{equation}
s_i = \alpha \, d_i + \beta \left(1 - \exp\!\left(-\lambda \frac{\Delta t_i}{3600}\right)\right),
\end{equation}
where $d_i = \mathrm{dist}(v_i, q)$ denotes the cosine distance between the memory embedding $v_i$ and the query embedding $q$, and $\Delta t_i$ represents the time difference between the current time and the memory timestamp. The parameter $\lambda > 0$ is a temporal decay coefficient that controls how strongly older observations are penalized.

Retrieval results can optionally be filtered by person identity $P$ and restricted to observations prior to time~$t_e$.

\item \textbf{Waypoint observations} $f(W, t_s, t_e)$:  
Retrieves all observations recorded at waypoint $W$ within the time interval $[t_s, t_e]$.

\item \textbf{Person observations} $f(P, t_s, t_e)$:  
Retrieves chronologically ordered observations associated with person identity $P$ within the time interval $[t_s, t_e]$.

\item \textbf{Persons per day} $f(d)$:  
Returns the set of unique person identities observed on date $d$.

\item \textbf{Daily summary for person} $f(d, P)$:  
Aggregates all observations of person $P$ on date $d$ and produces a LLM-generated summary of the individual’s activities.

\end{itemize}

The language model iteratively invokes these retrieval functions until it determines that sufficient evidence has been gathered to answer the query or the function-call limit is reached. This iterative reasoning process allows the model to progressively refine hypotheses about locations, time intervals, or activity patterns.

To enable embodied execution, the model is additionally provided with a control interface:

\begin{itemize}
\item \textbf{Navigate to waypoint} $f(W, P)$:
Commands the robot to navigate to waypoint $W$ for execution. 
Optionally, the function verifies whether person $P$ is present at the waypoint upon arrival.
\end{itemize}

This architecture integrates retrieval, reasoning, and action within a closed-loop framework. It enables the agent to leverage long-term experience, model temporal regularities in human behavior, estimate probable locations, and actively navigate in dynamic, human-centered environments.

\section{Data}

\textbf{COBD.} To test long-horizon human-centered reasoning, we collected the Collaborative Office Behavior Dataset (COBD), which contains navigation sequences recorded in a real-world indoor office environment using a Boston Dynamics Spot robot. During data acquisition, the robot traversed hallways, individual offices, and a shared kitchen, covering diverse functional spaces and yielding 136 distinct waypoints. 
Data were recorded at multiple times throughout the day over a period of 20 days, resulting in a total of 31 hours of robot operation. The dataset captures unconstrained and naturally occurring human activity. Individuals present in the environment followed their regular daily routines, and no scripted trajectories or staged interactions were imposed. Across recording days, participants appeared with varying clothing, hairstyles, and accessories, introducing realistic appearance changes over time. This setup ensures that the collected observations reflect realistic, temporally evolving human behavior in a dynamic workspace.

\begin{figure}[t]
\centering
\includegraphics[width=\linewidth]{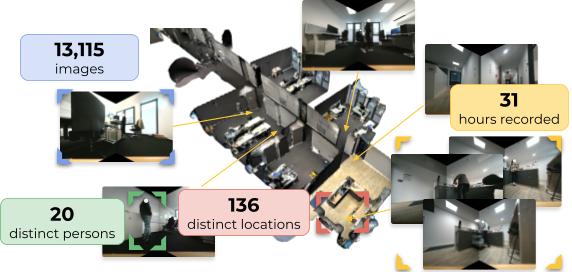}
\caption{The COBD dataset was collected over 20 days, with more than 31 hours of robot operation, yielding over 13,000 images across 136 distinct locations.}
\label{fig:summary_cobd}
\vspace{-4mm}
\end{figure}

Each sequence consists of temporally ordered waypoints synchronized with visual observations. The perception stream comprises stitched front-facing camera images, providing a continuous representation of the robot’s view during navigation. In addition to spatial and visual information, each waypoint is associated with timestamps, enabling long-horizon temporal analysis. Fig.~\ref{fig:summary_cobd} illustrates the dataset structure and summarizes the available data modalities.

\textbf{PersonEQA.} PersonEQA is a human-centered, long-horizon embodied question answering benchmark designed for navigation and video-based memory reasoning. It consists of temporally grounded and descriptive question–answer pairs derived from the COBD dataset and reformulated to reflect a persistent, multi-day memory setting. The benchmark evaluates a model’s ability to answer human-centered queries that require reasoning over extended temporal contexts and accumulated observations.
The benchmark comprises six question types, distributed as shown in Fig.~\ref{fig:summary_personeqa}. This distribution ensures coverage of both factual retrieval and higher-level reasoning tasks.

While some questions can be resolved through a single retrieval operation, others require aggregating observations across multiple days, identifying behavioral patterns, and performing temporal abstraction. Consequently, PersonEQA assesses not only memory retrieval accuracy but also long-horizon reasoning about human routines and interactions. 

%% file: content/04-experiments.tex
\section{Experiments}
We evaluate \papername on the PersonEQA benchmark under multiple configurations to analyze the influence of memory construction, caption generation, and language model selection on long-horizon reasoning performance.

\begin{figure}[t]
\centering
\includegraphics[width=\linewidth]{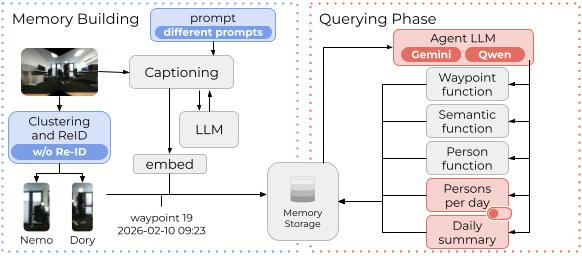}
\caption{We conduct ablation studies to evaluate individual component contributions. Components in gray remain fixed; the others are tested under varying configurations.}
\vspace{-4mm}
\label{fig:ablation}
\end{figure}

\textbf{Methods.}
\papername constructs a long-horizon memory consisting of clustered person identities and semantically enriched scene captions. To examine the impact of caption design, we evaluate multiple prompting strategies for caption generation and analyze their influence on downstream retrieval and reasoning performance (see~Fig.~\ref{fig:ablation}). 

Building on this memory representation, the system provides up to five structured retrieval functions that allow targeted access to semantic observations, waypoint histories, and person-specific information. To assess the contribution of hierarchical memory access, we conduct ablation studies in which only three core retrieval functions (person history, waypoint history, and semantic observations) are available to the LLM. In addition, we vary the maximum number of allowed retrieval function calls per query to analyze how depth of reasoning affects performance. These experiments quantify the benefit of higher-level aggregation mechanisms such as person summaries and available-person queries.

The retrieved memory subset is subsequently passed to a language model for reasoning. We evaluate a closed-source model (Gemini 3 Flash) and multiple open-source Qwen VLMs  \cite{Qwen-VL} at different scales to assess robustness across reasoning backbones. For Gemini, the context window is fixed to 1{,}000{,}000 tokens, whereas Qwen models operate with a 262{,}144-token limit. All configurations are compared against a full-context baseline that processes the complete caption log up to the query timestamp without structured retrieval. Because the goal is to support long-term reasoning within a persistent environment, evaluation uses memory collected in the same site in which questions are asked.

\textbf{Metrics.}
PersonEQA comprises six question categories, each evaluated with task-specific criteria:

\begin{itemize}
\item \textbf{Spatial questions:} The model predicts a waypoint in 3D space. Performance is measured by Euclidean distance to the ground-truth waypoint, with predictions within 5~m considered correct.

\item \textbf{Temporal questions (Duration and Time):} The model outputs a date and time, while duration questions require an answer in minutes. Predictions within a tolerance of $\pm$5 minutes are counted as correct.

\item \textbf{Binary questions:} The model outputs either \textit{yes} or \textit{no}. Accuracy is computed via exact label matching.

\item \textbf{Descriptive questions:} The model generates free-form text responses. Answers are evaluated using an LLM-based grading protocol similar to Anwar et al.~\cite{anwar_remembr_2024}. A response is considered correct if it contains the key semantic elements of the ground truth.

\item \textbf{Person identification questions:} The model outputs one or multiple person names. A prediction is counted as correct if the predicted identity set exactly matches the ground-truth set.
\end{itemize}

All \papername results are averaged over three random seeds. Due to computational constraints, the baseline is evaluated using a single seed. We froze the test set before evaluation and made no changes after testing began.

\begin{figure}[t]
\centering
\includegraphics[width=\linewidth]{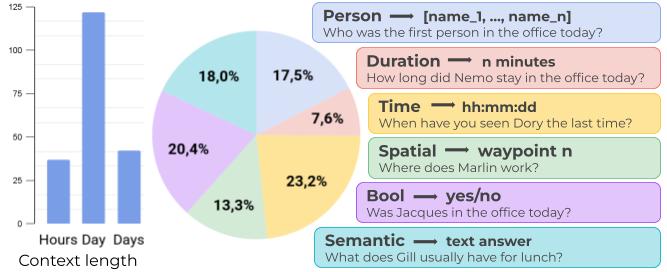}
\caption{PersonEQA comprises 200 questions across six answer types. The required context provided to the LLM may span a few hours, a single day, or multiple days.}
\vspace{-2mm}
\label{fig:summary_personeqa}
\end{figure}

\begin{table*}[t]
\setlength{\tabcolsep}{3pt}
\centering
\caption{Comparison of \papername{} with different backbone LLMs against a full-context baseline using Gemini 3 Flash.  The baseline and the \papername{} (Gemini 3 Flash) entry use identical prompting, temperature, and decoding parameters.}
\label{tab:results_baseline}
\scalebox{0.91}{
\begin{tabular}{llcccccccc}
\toprule
 & & \multicolumn{7}{c}{\textbf{Accuracy (in \%)}} & \\
&
 & Total 
 & Person 
 & Bool 
 & Time 
 & Semantic 
 & Spatial 
 & Duration 
 & Token/Quest.\\
\midrule
Ours & Qwen3-VL 235B 
& $66.01 \pm 1.03$ & $57.14 \pm 2.86$ & $71.32 \pm 1.34$ & $63.04 \pm 2.18$ & $59.34 \pm 1.75$ & $78.21 \pm 5.88$ & $75.00 \pm 0.00$ & 11{,}589 \\

& Qwen3-VL-235B (–ReID) 
& $60.49 \pm 1.15$ & $48.57 \pm 0.00$ & $74.21 \pm 2.35$& $60.15 \pm 2.51$ & $56.48 \pm 6.41$ & $60.25 \pm 4.44$ & $60.42 \pm 9.55$ & 11{,}790 \\

& Qwen3-VL-235B (3 func) 
& $63.04 \pm 0.75$ & $44.76 \pm 1.65$ & $70.54 \pm 3.55$ & $60.15 \pm 1.25$ & $51.85 \pm 1.61$ & $84.62 \pm 3.85$ & $\mathbf{81.25 \pm 0.00}$ & 12{,}976 \\

& Qwen3-235B Thinking
& $63.20 \pm 0.29$ & $48.57 \pm 2.86$ & $62.02 \pm 2.68$ & $60.15 \pm 1.25$ & $63.89 \pm 4.82$ & $\mathbf{95.90 \pm 2.22}$ & $68.75 \pm 6.25$ & 12{,}415 \\

& Qwen3-30B Thinking
& $57.92 \pm 2.62$ & $47.62 \pm 1.65$ & $60.47 \pm 0.00$ & $53.61 \pm 1.24$ & $50.93 \pm 9.75$ & $79.49 \pm 2.22$ & $66.67 \pm 9.55$ & \textbf{9{,}430} \\

& Gemini 3 Flash
& $\mathbf{75.41 \pm 0.28}$ & $60.95 \pm 1.65$ & $\mathbf{81.40 \pm 0.00}$ & $65.94 \pm 3.32$ & $\mathbf{81.48 \pm 1.60}$ & $92.31 \pm 3.85$ & $77.08 \pm 7.22$ & 106{,}347\\

\hline
Baseline & All captions in context 
& 67.33 & \textbf{65.71} & \textbf{81.40} & \textbf{67.39} & 75.00 & 34.62 & 68.75 & 631{,}651 \\
 \bottomrule
\end{tabular}
}
\vspace{-2mm}
\end{table*}

\begin{table}[t]
\setlength{\tabcolsep}{3pt}
\centering
\caption{Function Call Limits}
\label{tab:results_btool_call_limit}
\scalebox{0.8}{
\begin{tabular}{lcccccccc}
\toprule
 & & \multicolumn{7}{c}{\textbf{Accuracy (in \%)}}  \\
& Limit
 & Total 
 & Person 
 & Bool 
 & Time 
 & Semantic 
 & Spatial 
 & Duration \\
\midrule
Qwen3-VL-235B & 5 
& 62.29 & 51.43 &69.77 & 64.49 & 62.96 & 76.62 & 72.67 \\

&10
& 66.01 & 57.14 & 71.32 & 63.04 & 59.34 & 78.21 & 75.00\\ 

& 15
& 65.07 & 52.38 & 71.32 & 62.32 & 62.59 & 74.00 & 75.00\\ 

\hline
Gemini 3 Flash & 5
& 73.88 & 60.50 & 72.95 & 63.77 & \textbf{82.41} & 91.03 & 75.00\\ 

&10
& \textbf{75.41} & 60.95 & 81.40 & \textbf{65.95} & 81.48 & \textbf{92.31} & 77.08\\ 

& 15
& 74.59 & \textbf{62.86} & \textbf{83.72} & 65.22 & 72.22 & 91.03 & \textbf{81.25}\\ 

 \bottomrule
\end{tabular}
}
\end{table}

\begin{table}[t]
\centering
\caption{Caption Prompt Evaluation}
\label{tab:caption_results}
\scalebox{0.75}{
\begin{tabular}{lccccccc}
\toprule
 & \multicolumn{7}{c}{\textbf{Accuracy (in \%)}} \\
 & Total 
 & Person 
 & Bool 
 & Time 
 & Semantic 
 & Spatial 
 & Duration\\
\midrule
Hand-focused 
& 63.52 & 51.43 & \textbf{73.26} & 59.42 & 58.34 & 72.67 & 72.92 \\

Interaction-centered 
& \textbf{66.01} & \textbf{57.14} & 71.32 & \textbf{63.04} & 59.34 & \textbf{78.21} & 75.00 \\

Movement-aware 
& 64.69 & 50.48 & 71.32 & 59.42 & \textbf{61.11} & 76.92 & \textbf{81.25}\\
\bottomrule
\end{tabular}
}
\vspace{-4mm}
\end{table}

%% file: content/05-results.tex
\begin{table*}[t]
\centering
\caption{Real World Evaluation}
\label{tab:real_world_results}
\setlength{\tabcolsep}{5pt}
\scalebox{0.98}{
\begin{tabular}{lcccccccc}
\toprule
 & Category
 & \makecell{Success\\Rate (\%)}
 & \makecell{Duration\\(min)}
 & \makecell{Waypoints\\Visited}
 & \makecell{Function\\Calls}
 & \makecell{First-Visit\\Detection (\%)}
 & \makecell{Direct\\Path (\%)}
 & Tokens \\
\midrule
Find Dory & a & $90$ & $3.66 \pm 1.44$ & $5.4 \pm 2.27$ & $0.9 \pm 0.32$ & 90 & 80 & 192{,}510 \\

Are Early Birds already on-site & a & $60$ & $6.25 \pm 1.56$ & $5.5 \pm 0.67$ & $8.5 \pm 4.32$ & $60$ & 100 & 1{,}434{,}385  \\

Pick up visitor and show around (Fig~\ref{fig:live_nala}) & b & 100 & $2.31 \pm 0.59$ & $3.2 \pm 0.63$ & $1.8 \pm 0.79$& 100 & 90  & 297{,}808 \\

Visit people and give summary & b & 80 & $2.83 \pm 0.79$ & $3.14 \pm 0.31$ & $1.57 \pm 0.64$ & 70 & 100 & 87{,}997 \\

Pick up package and bring back (Fig~\ref{fig:live_package}) & c & 100 & $2.24 \pm 1.27$ & $3.33 \pm 1.58$ & $4.11 \pm 2.20$ & 100 & 80 & 425{,}139 \\

Find people in partial occlusion & d & 50 & $3.31 \pm 1.24$ & $5.8 \pm 2.49$ & $11.8 \pm 7.1$ & 30 & 70 & 581{,}085 \\
\bottomrule
\end{tabular}
}
\vspace{0mm}
\end{table*}

\section{Results}
\begin{figure}[t]
\centering
\begin{tikzpicture}
\begin{axis}[
    width=0.9\linewidth,
    height=5.5cm,
    xlabel={Tokens per Question},
    ylabel={Accuracy (\%)},
    xmin=0,
    xmax=800000,
    ymin=55,
    ymax=80,
    xmode=log,
    scaled x ticks=false,
    grid=both,
    grid style={dashed,gray!20},
    tick label style={font=\small},
    label style={font=\small},
    legend style={
        font=\scriptsize,
        at={(axis description cs:0.98,0.02)}, 
        anchor=south east,
        draw=none,
        fill=white,
        align=left,
        cells={anchor=west},
        inner sep=3pt,
        row sep=1pt,
        nodes={inner sep=0pt},
    },
]

\addplot[
    only marks,
    mark=*,
    mark size=3.5pt,
    color=blue_google
] coordinates {(11589,66.01)};
\addlegendentry{Qwen3-VL-235B}

\addplot[
    only marks,
    mark=*,
    mark size=3.5pt,
    color=red_google
] coordinates {(12415,63.20)};
\addlegendentry{Qwen3-235B Thinking}

\addplot[
    only marks,
    mark=*,
    mark size=3.5pt,
    color=yellow_google
] coordinates {(9430,57.92)};
\addlegendentry{Qwen3-30B Thinking}

\addplot[
    only marks,
    mark=*,
    mark size=3.5pt,
    color=green_google
] coordinates {(106347,75.41)};
\addlegendentry{Gemini 3 Flash}

\addplot[
    only marks,
    mark=*,
    mark size=3.5pt,
    color=pink_google
] coordinates {(631651,67.33)};
\addlegendentry{Baseline}

\end{axis}
\end{tikzpicture}
\caption{\textbf{Accuracy versus token usage (log-scale).} Leveraging data retrieval and employing Gemini as the agent LLM yields superior performance while consuming only 17\% of the tokens. In contrast, using Qwen achieves performance comparable to the baseline while requiring merely 2\% of the tokens. See Table~\ref{tab:results_baseline} for numerical results.}
\label{fig:accuracy_plot}
\vspace{-4mm}
\end{figure}
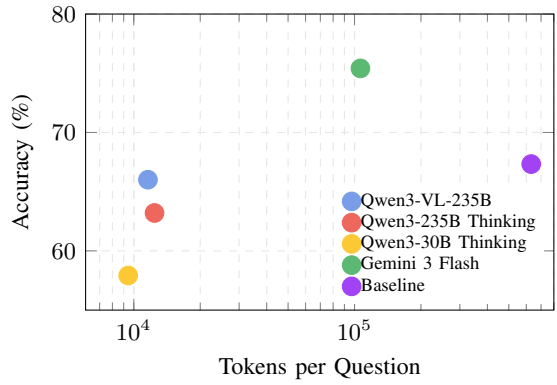

\textbf{\papername achieves higher accuracy than the baseline while using 83\% fewer tokens.}
As shown in Fig.~\ref{fig:accuracy_plot}, structured retrieval reduces tokens, cutting computational overhead and latency while improving performance. \papername{} with Gemini 3 Flash excels in spatial, semantic, and duration reasoning, critical for embodied navigation. The baseline slightly outperforms \papername{} on person identification and time-of-event questions, likely reflecting exhaustive access to all captions and timestamps, facilitating lookup. \papername{} substantially improves spatial reasoning, suggesting structured retrieval mitigates context dilution.

\textbf{\papername in the open-source setup matches baseline accuracy while using only 2\% of the tokens.}
Fig.~\ref{fig:accuracy_plot} shows that the open-source configuration of \papername achieves a substantial reduction in token usage, leading to lower computational cost and reduced inference latency, without degrading performance. Furthermore, the open-source model can be self-hosted, enabling deployment without reliance on closed-source APIs.

\textbf{Context Dilution Appears in the Full-Context Baseline.}
Spatial reasoning over large, unstructured contexts suffers from attention diffusion across repetitive waypoint identifiers. Although the full memory log provides complete historical access, it reduces the signal-to-noise ratio. Because every entry contains a waypoint reference, the model must select among many similar spatial tokens, increasing ambiguity and weakening spatial grounding.

\textbf{Increasing the function-call limit does not necessarily improve performance.} 
As shown in Table~\ref{tab:results_btool_call_limit}, neither higher nor lower limits on the number of retrieval function calls consistently lead to improved accuracy. Empirically, a limit of ten function calls appears to provide an effective balance, allowing the model to gather sufficient relevant information while avoiding the inclusion of extraneous context.

\begin{figure*}[t]
\centering
\includegraphics[width=\linewidth]{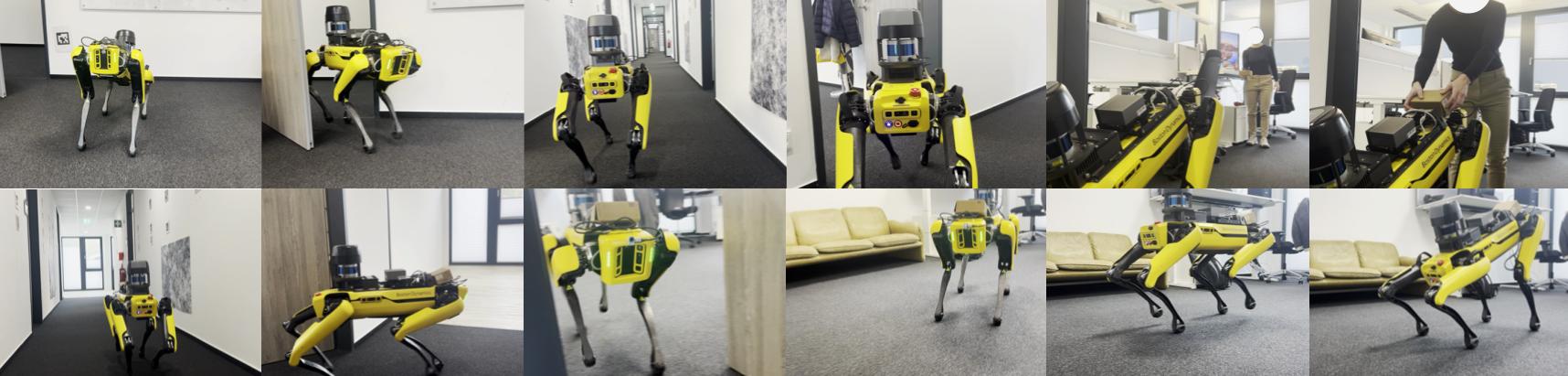}
\caption{Package: “Please go to Dory's workplace and pick up a package that she has for me. When you have received the package come back to me.“}
\label{fig:live_package}
\vspace{2mm}
\centering
\includegraphics[width=\linewidth]{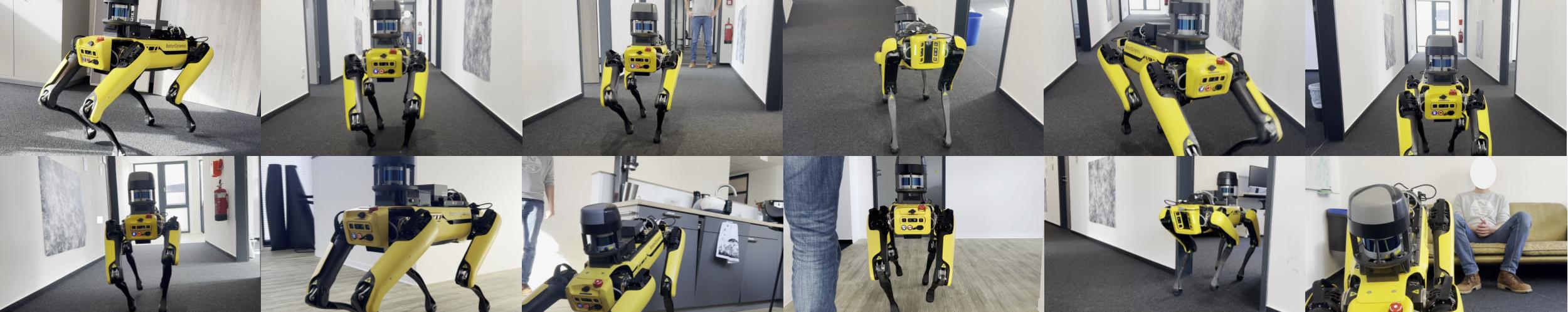}
\caption{Visitor: “Pick up our visitor at waypoint 168. Then bring him to the kitchen, and after arriving, show him where he can sit on a couch.“}
\label{fig:live_nala}
\vspace{2mm}
\centering
\includegraphics[width=\linewidth]{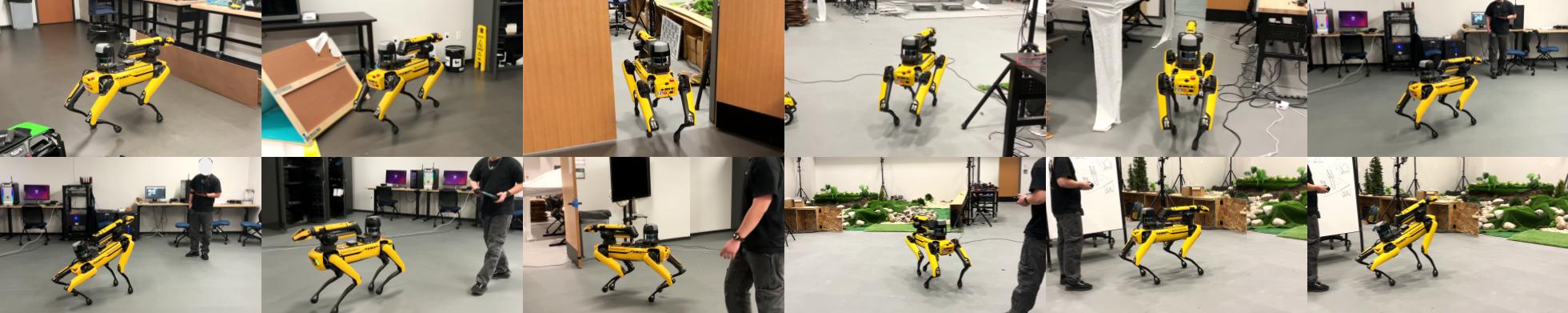}
\caption{“Find Nemo and bring him to the outdoor scene setup“}
\label{fig:live_lunch}
\vspace{0mm}
\end{figure*}

\section{Real-World Deployment}
Although PersonEQA provides a controlled benchmark for long-horizon reasoning, it does not capture the practical constraints of real-time robotic deployment. 

\textbf{Robot Deployment.} To evaluate performance under realistic conditions, we deploy \papername on a Boston Dynamics Spot robot. The deployment is conducted in the same office environment used for collecting the COBD dataset. Historical COBD recordings are preloaded into the robot’s long-term memory, enabling reasoning over previously observed human routines while operating in the live environment. During evaluation, the robot is assigned navigation and person-search tasks that require both memory-based inference and embodied execution. Low-level navigation is handled by the Boston Dynamics GraphNav stack \cite{BD_GraphNavPythonClient}, whereas high-level reasoning and waypoint selection are generated by \papername.

We evaluate the system on six representative tasks, each repeated over ten runs. These tasks can be grouped into the following categories:

\textbf{(a) Routine-Based Search:} Require leveraging accumulated behavioral patterns to predict likely person locations.

\textbf{(b) Multi-Step Task Execution:} Involving sequential reasoning and navigation across multiple subtasks and locations.

\textbf{(c) Person–Object Interaction and Delivery:} Combining person identification, object detection, and goal-directed navigation.

\textbf{(d) Perception Robustness:} Assesses face detection and person re-identification under occlusion and difficult lighting, demanding reasoning within the perception–action loop.

For the real-world experiments, we employ Gemini 3 Flash as the reasoning model with the full set of retrieval functions and the best-performing caption configuration (see Table~\ref{tab:caption_results}).

\textbf{Qualitative results.}
The results in Table~\ref{tab:real_world_results} demonstrate that the system performs reliably across diverse real-world tasks.
The number of visited waypoints and the relatively low number of function calls indicate that navigation was typically guided by accurate routine-based predictions rather than exhaustive search. For highly complex queries, function calls and token usage increase, while navigation remains efficient, requiring only a few visited waypoints. Failures primarily occurred in situations where the target person was partially occluded (e.g., by a chair), observed from the back, or affected by strong backlighting or low illumination. In these cases, face detection and person ReID were less reliable, leading to reduced \emph{First Occurrence} rates and occasional search errors. Also, when the target person is absent, the model may hallucinate prior observations of that individual. Overall, experiments across both environments show that long-horizon routine modeling enables efficient and goal-directed navigation under varying spatial layouts and crowd conditions, while perception robustness particularly under challenging visual conditions remains the primary limitation.

%% file: content/privacy.tex
\section{Ethics and Privacy}
This system models persistent human routines and thus raises privacy risks. Data were collected with participant consent and withdrawal rights under appropriate ethics oversight only for the minimum period needed for research. We store pseudonymous IDs and derived memory representations, restrict access to raw logs, and exclude personally identifying information from released materials. We do not support non-consensual tracking or surveillance. Outputs are probabilistic and should not be used for high-stakes decisions.

%% file: content/06-conclusions.tex
\section{Conclusions}
We introduced \papername, a framework for robot navigation grounded in long-horizon, human-centered memory. The system is designed to operate in highly dynamic environments and explicitly models recurring human routines to inform decision-making and predictive embodied navigation. Our experimental results demonstrate that \papername enables a robot to answer questions about individuals’ activities and daily habits, and to execute person-search tasks across two distinct real-world environments in a structured and memory-informed manner. By leveraging targeted retrieval over accumulated observations, the system improves answer accuracy while substantially reducing the number of processed tokens compared to non-retrieval baselines. 

\textbf{Limitations and Future Work}. While our memory log is generated from the robot’s camera stream, it does not incorporate additional spatial or personal information (e.g., room labels). Incorporating such information could enhance the richness of the memory representation and provide additional utility for downstream tasks. Moreover, the current LLM-based agent could be replaced or complemented with a VLM capable of directly processing visual input, such as recent images, rather than relying solely on captions, thereby providing a broader and more detailed scene representation. Although our results are presently comparable to the baseline, the importance of efficient data retrieval is expected to increase as the volume of visual data grows. Consequently, developing methods for scalable and effective memory aggregation represents a critical direction for future~research.

%% file: content/acknowledgement.tex
\section*{Acknowledgement}
The authors thank the anonymous reviewers for their valuable
feedback. The research was conducted under the RobOdin project funded by the European Union through it's Interreg Germany-Denmark funding program. We acknowledge technical support by the Embodied AI Center at Kiel University.

%% file: root.bbl
\begin{thebibliography}{10}
\providecommand{\url}[1]{#1}
\csname url@samestyle\endcsname
\providecommand{\newblock}{\relax}
\providecommand{\bibinfo}[2]{#2}
\providecommand{\BIBentrySTDinterwordspacing}{\spaceskip=0pt\relax}
\providecommand{\BIBentryALTinterwordstretchfactor}{4}
\providecommand{\BIBentryALTinterwordspacing}{\spaceskip=\fontdimen2\font plus
\BIBentryALTinterwordstretchfactor\fontdimen3\font minus \fontdimen4\font\relax}
\providecommand{\BIBforeignlanguage}[2]{{%
\expandafter\ifx\csname l@#1\endcsname\relax
\typeout{** WARNING: IEEEtran.bst: No hyphenation pattern has been}%
\typeout{** loaded for the language `#1'. Using the pattern for}%
\typeout{** the default language instead.}%
\else
\language=\csname l@#1\endcsname
\fi
#2}}
\providecommand{\BIBdecl}{\relax}
\BIBdecl

\bibitem{Cadena_2016}
C.~Cadena, L.~Carlone, H.~Carrillo, Y.~Latif, D.~Scaramuzza, J.~Neira, I.~Reid, and J.~J. Leonard, ``{Past, Present, and Future of Simultaneous Localization and Mapping: Toward the Robust-Perception Age},'' \emph{IEEE Transactions on Robotics}, vol.~32, no.~6, pp. 1309--1332, 2016.

\bibitem{Armeni20193DSG}
I.~Armeni, Z.-Y. He, J.~Gwak, A.~Zamir, M.~Fischer, J.~Malik, and S.~Savarese, ``{3D Scene Graph: A Structure for Unified Semantics, 3D Space, and Camera},'' in \emph{ICCV}, 2019, pp. 5663--5672.

\bibitem{Werby-RSS-24}
A.~Werby, C.~Huang, M.~Büchner, A.~Valada, and W.~Burgard, ``{Hierarchical Open-Vocabulary 3D Scene Graphs for Language-Grounded Robot Navigation},'' in \emph{RSS}, 2024.

\bibitem{doi:10.1177/02783649211056674}
A.~Rosinol, A.~Violette, M.~Abate, N.~Hughes, Y.~Chang, J.~Shi, A.~Gupta, and L.~Carlone, ``{Kimera: From SLAM to Spatial Perception with 3D Dynamic Scene Graphs},'' \emph{The International Journal of Robotics Research}, vol.~40, no. 12--14, pp. 1510--1546, 2021.

\bibitem{chiang_mobility_2024}
Z.~Xu \emph{et~al.}, ``Mobility {VLA}: {Multimodal} {Instruction} {Navigation} with {Long}-{Context} {VLM}s and {Topological} {Graphs},'' in \emph{CoRL}, ser. Proceedings of Machine Learning Research, vol. 270, 2025, pp. 3866--3887.

\bibitem{ahn_as_2022}
M.~Ahn \emph{et~al.}, ``Do {As} {I} {Can} and {Not} {As} {I} {Say}: Grounding language in robotic affordances,'' in \emph{CoRL}, 2022.

\bibitem{shah_lm-nav_2022}
D.~Shah, B.~Osi\'nski, B.~Ichter, and S.~Levine, ``{LM-Nav}: {Robotic Navigation with Large Pre-Trained Models of Language, Vision, and Action},'' in \emph{CoRL}, ser. Proceedings of Machine Learning Research, vol. 205, 2023, pp. 492--504.

\bibitem{espada_leveraging_2025}
J.~P. Espada, S.~Y. Qiu, R.~G. Crespo, and J.~L. Carús, ``{Leveraging Large Language Models for Autonomous Robotic Mapping and Navigation},'' \emph{International Journal of Advanced Robotic Systems}, vol.~22, no.~2, p. 17298806251325965, 2025.

\bibitem{bian_large_2025}
S.~Bian, Y.~Zhang, G.~Tian, Z.~Miao, E.~Q. Wu, S.~X. Yang, and C.~Hua, ``Large {Language} {Model}-{Based} {Task} {Planning} for {Service} {Robots}: A {Review},'' \emph{Biomimetic Intelligence and Robotics}, vol.~6, no.~1, p. 100274, 2026.

\bibitem{park2023generativeagentsinteractivesimulacra}
J.~S. Park, J.~O'Brien, C.~J. Cai, M.~R. Morris, P.~Liang, and M.~S. Bernstein, ``{Generative Agents: Interactive Simulacra of Human Behavior},'' in \emph{UIST}, 2023.

\bibitem{ginting2025entermindpalacereasoning}
M.~F. Ginting \emph{et~al.}, ``{Enter the Mind Palace: Reasoning and Planning for Long-term Active Embodied Question Answering},'' in \emph{CoRL}, 2025.

\bibitem{anwar_remembr_2024}
A.~Anwar, J.~Welsh, J.~Biswas, S.~Pouya, and Y.~Chang, ``{ReMEmbR: Building and Reasoning Over Long-Horizon Spatio-Temporal Memory for Robot Navigation},'' in \emph{ICRA}, 2025, pp. 2838--2845.

\bibitem{bostondynamics_spot}
{Boston Dynamics}, ``{Spot -- Agile mobile robot},'' \url{https://bostondynamics.com/}, accessed: Feb. 14, 2026.

\bibitem{majumdar_openeqa_nodate}
A.~Majumdar \emph{et~al.}, ``{OpenEQA}: Embodied {Question} {Answering} in the {Era} of {Foundation} {Models},'' in \emph{CVPR}, 2024, pp. 16\,488--16\,498.

\bibitem{ren_explore_2024}
A.~Z. Ren, J.~Clark, A.~Dixit, M.~Itkina, A.~Majumdar, and D.~Sadigh, ``{Explore until Confident: Efficient Exploration for Embodied Question Answering},'' in \emph{RSS}, 2024.

\bibitem{zhao_cityeqa_2025}
Y.~Zhao, K.~Xu, Z.~Zhu, Y.~Hu, Z.~Zheng, Y.~Chen, Y.~Ji, C.~Gao, Y.~Li, and J.~Huang, ``{CityEQA}: A {Hierarchical} {LLM} {Agent} on {Embodied} {Question} {Answering} {Benchmark} in {City} {Space},'' 2025, arXiv:2502.12532.

\bibitem{grauman_ego4d_2022}
kristen grauman \emph{et~al.}, ``{ego4d: around the world in 3,000 hours of egocentric video},'' in \emph{cvpr}, 2022.

\bibitem{zhou_navgpt_2023}
G.~Zhou, Y.~Hong, and Q.~Wu, ``{NavGPT: explicit reasoning in vision-and-language navigation with large language models},'' in \emph{AAAI}, 2024.

\bibitem{yadav_findingdory_2025}
K.~Yadav, Y.~Ali, G.~Gupta, Y.~Gal, and Z.~Kira, ``{FindingDory: A Benchmark to Evaluate Memory in Embodied Agents},'' 2025, arXiv:2506.15635.

\bibitem{singh_progprompt_2022}
I.~Singh, V.~Blukis, A.~Mousavian, A.~Goyal, D.~Xu, J.~Tremblay, D.~Fox, J.~Thomason, and A.~Garg, ``{ProgPrompt: Generating Situated Robot Task Plans using Large Language Models},'' in \emph{ICRA}, 2023, pp. 11\,523--11\,530.

\bibitem{sun_interactive_2023}
L.~Sun, D.~K. Jha, C.~Hori, S.~Jain, R.~Corcodel, X.~Zhu, M.~Tomizuka, and D.~Romeres, ``{Interactive Planning Using Large Language Models for Partially Observable Robotics Tasks},'' in \emph{ICRA}, 2024, pp. 14\,054--14\,061.

\bibitem{Somers_2024}
V.~Somers, A.~Alahi, and C.~D. Vleeschouwer, ``{Keypoint Promptable Re-Identification},'' in \emph{ECCV}, ser. Lecture Notes in Computer Science, vol. 15137, 2024, pp. 216--233.

\bibitem{he2025instructreidmultipurposepersonreidentification}
W.~He \emph{et~al.}, ``{Instruct-ReID}: {A Multi-Purpose Person Re-Identification Task with Instructions},'' in \emph{CVPR}, 2023, pp. 17\,521--17\,531.

\bibitem{liang2025differdisentanglingidentityfeatures}
X.~Liang and Y.~S. Rawat, ``{DIFFER: Disentangling Identity Features via Semantic Cues for Clothes-Changing Person Re-ID},'' in \emph{CVPR}, June 2025, pp. 13\,980--13\,989.

\bibitem{pathak2025colorscolorsignoreclothes}
P.~Pathak and Y.~S. Rawat, ``{Colors See Colors Ignore: Clothes Changing ReID with Color Disentanglement},'' in \emph{ICCV}, 2025, pp. 16\,797--16\,807.

\bibitem{he2016deep}
K.~He, X.~Zhang, S.~Ren, and J.~Sun, ``{Deep residual learning for image recognition},'' in \emph{CVPR}, 2016, pp. 770--778.

\bibitem{BD_GraphNavPythonClient}
\BIBentryALTinterwordspacing
{Boston Dynamics, Inc.}, ``{Graph Nav (bosdyn.client.graph\_nav)},'' Spot SDK Python API Reference (v5.1.1), 2026, accessed: 2026-02-18. [Online]. Available: \url{https://dev.bostondynamics.com/python/bosdyn-client/src/bosdyn/client/graph\_nav.html}
\BIBentrySTDinterwordspacing

\bibitem{Qwen-VL}
J.~Bai, S.~Bai, S.~Yang, S.~Wang, S.~Tan, P.~Wang, J.~Lin, C.~Zhou, and J.~Zhou, ``{Qwen-VL}: {A Versatile Vision-Language Model for Understanding, Localization, Text Reading, and Beyond},'' 2023, arXiv:2308.12966.

\bibitem{emb2024mxbai}
\BIBentryALTinterwordspacing
S.~Lee, A.~Shakir, D.~Koenig, and J.~Lipp, ``{Open Source Strikes Bread - New Fluffy Embeddings Model},'' 2024. [Online]. Available: \url{https://www.mixedbread.ai/blog/mxbai-embed-large-v1}
\BIBentrySTDinterwordspacing

\bibitem{ultralytics_2024}
\BIBentryALTinterwordspacing
{Ultralytics}, ``{Ultralytics YOLO11-Pose},'' 2024, software repository. [Online]. Available: \url{https://github.com/ultralytics/ultralytics}
\BIBentrySTDinterwordspacing

\bibitem{insightface2023github}
I.~Contributors, ``{InsightFace: 2D and 3D Face Analysis Project},'' \url{https://github.com/deepinsight/insightface}, 2023.

\bibitem{ester1996dbscan}
M.~Ester, H.-P. Kriegel, J.~Sander, and X.~Xu, ``{A density-based algorithm for discovering clusters in large spatial databases with noise},'' in \emph{KDD}, 1996, pp. 226--231.

\end{thebibliography}
